# Towards Stable Adversarial Feature Learning for LiDAR based Loop Closure Detection *

1st Lingyun Xu
*State Key Laboratory of Robotics*
*Chinese Academy of Sciences*
Shenyang, China
xulingyun@sia.cn

2nd Peng Yin
*State Key Laboratory of Robotics*
*Chinese Academy of Sciences*
Shenyang, China
yinpeng@sia.cn

3rd Haibo Luo
*Key Laboratory of Optical-Electronics Information Processing*
*Chinese Academy of Sciences*
Shenyang, China
luohb@sia.cn

4th Yunhui Liu
*William M.W. Mong Engineering Building*
*Chinese University of Hong Kong*
Hong Kong, China
yunhui.liu@gmail.com

5th Jianda Han
*State Key Laboratory of Robotics*
*Chinese Academy of Sciences*
Shenyang, China
jdhan@sia.cn

*Abstract*—Stable feature extraction is the key for the Loop closure detection (LCD) task in the simultaneously localization and mapping (SLAM) framework. In our paper, the feature extraction is operated by using a generative adversarial networks (GANs) based unsupervised learning. GANs are powerful generative models, however, GANs based adversarial learning suffers from training instability. We find that the data-code joint distribution in the adversarial learning is a more complex manifold than in the original GANs. And the loss function that drive the attractive force between synthesis and target distributions is unable for efficient latent code learning for LCD task. To relieve this problem, we combines the original adversarial learning with an inner cycle restriction module and a side updating module. To our best knowledge, we are the first to extract the adversarial features from the light detection and ranging (LiDAR) based inputs, which is invariant to the changes caused by illumination and appearance as in the visual inputs. We use the KITTI odometry datasets to investigate the performance of our method. The extensive experiments results shows that, with the same LiDAR projection maps, the proposed features are more stable in training, and could significantly improve the robustness on viewpoints differences than other state-of-art methods.

## I. Introduction

Loop closure detection (LCD) is the essential module in the simultaneously localization and mapping (SLAM) task. Traditionally, SLAM could be divided into two categories, metric SLAM [1,2] and Appearance (or Topological) SLAM [3], where the former could achieve accurate localization and mapping results but need huge computation power and storage requirement. For the long term large scale navigation task, such as highway traveling [4], or life-long navigation [5] task, metric based SLAM couldn't satisfy the real time requirement on the current normal mobile robots. In contract, the features used in appearance based SLAM methods usually in low dimension formation and are easy to store for long term navigation task, such as in FABMAP [6]. Feature extraction is the key in the appearance based SLAM, and the desired features should capture the major geometry of the local scene and ignore the condition (illumination, appearance, viewpoints) changes at the same time.

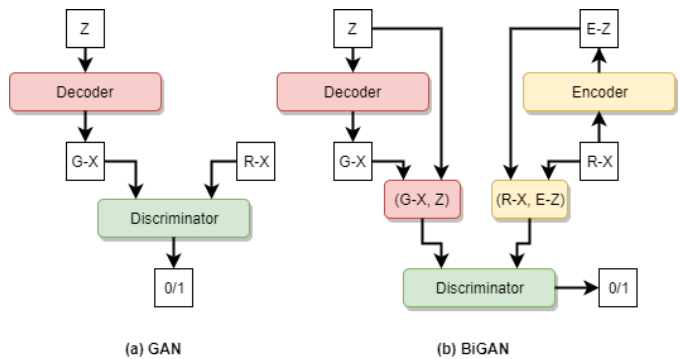

**Figure 1** The structure of DCGAN and BiGAN.

Traditional appearance based SLAM, such as FABMAP [3,6], RatSLAM [7] and SeqSLAM [8] [4], usually rely on handcraft features [9–12] for LCD task. For visual inputs, such features couldn't avoid the illumination changes from data to night, and appearance changes from weather-to-weather or season-to-season. Comparing with the visual inputs, LiDAR based point-clouds is inherently invariant to the illumination and appearance changes. In this paper, we use the 2D top-view maps extracted from the LiDAR inputs to represent the local scenes. In application, the LCD accuracy is highly relies on the feature extraction step. Since the traditional handcraft features [9,10,12] could not capture the local detail and the global connections at the same time, so in the big viewpoints case, such features may fail to achieve the accurate LCD.

With the development in deep neural networks (DNN), some researchers tried to investigate the LCD ability with DNN features. Sunderhauf [13] et.al investigate the ability of each layer in different ConvNets for LCD. Lowry [14] proposed the PCA based common feature eliminating for appearance invariant feature extraction. Most recently, Chen [15] proposed a simple training method to enable the feature training for LCD task, where they divide the place scene into several classes. The above methods could only be used in visual inputs, and must be supported with data labels [16].

To enable the feature extraction from the unlabeled 2D top-view LiDAR maps, we use an unsupervised learning method, bidirectional generative adversarial networks (BiGAN) [17]. BiGANs is an adversarial learning version of the generative adversarial networks (GANs) [18], where GANs is a kind of unsupervised generative models as shown in Figure 1(a). The original GANs could not inference the latent code from data domain, and is combined with a decoder module to generate synthesis data from random noise, and a discriminator module to distinguish the real data from the synthesis ones. To enable the code inference, BiGANs adds an additional encoder module to mapping the real data into the latent code domain, and the discriminator is also updated to distinguish the data-code joint distribution. BiGANs has strong ability in data generalization, and could extraction efficient features from unseen images with only limited training samples.

However, BiGANs itself is hard to train than the original GANs. We find that the data-code joint distribution in the adversarial learning is a more complex manifold than in the original GANs. As prove by Arjovsky et.al [19], when measures with Jensen-Shannon Divergence (JSD), the distance of the two complex distribution is "equal to constant at almost anywhere", i.e. the gradient for the networks updating is equal to 0 almost all the time. Besides, the loss function that driven the attractive force between synthesis and target distributions is unstable for efficient latent code learning for the LCD task.

In this paper, we proposed a stable adversarial feature learning (Stable-AFL) method for LiDAR based loop closure detection. A natural thought in our work is to enhance the attractive forces between the joint distributions, and at the same time balance the convergences in both data and code domains to avoid excessive forces harming the feature uniqueness. To achieve such goal, we combines the original adversarial learning with an inner cycle restriction module and a side updating module. The major contributions of this paper could be concluded as:

- We propose an unsupervised feature learning framework for LiDAR projection map based LCD. The LiDAR inputs are invariant to illumination and appearance changes. Our method could extract efficient features from the 2D top-view maps in real time, and could be easily embedded into normal mobile robots.

- We proposed a stable adversarial feature learning method, which is based on the original BiGAN, and combined with the cycle restriction module and the side updating module. The proposed method could enhance the attractive force between the joint distributions, and assist the efficient feature extraction for the LCD task.

The rest of this paper is organized as following: in Section 2, we introduce the BiGAN based adversarial learning for LCD; Section 3 is the major part to explain our stable adversarial feature learning method; Section 4 demonstrates the results of our method on KITTI odometry benchmark; finally, Section 5 shows the conclusion and future consideration.

## II. ADVERSARIAL LEARNING BASED LCD

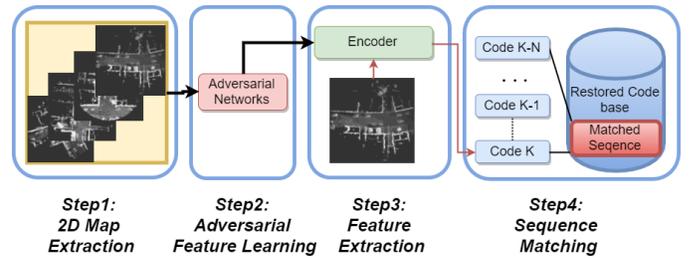

**Figure 2** The Adversarial Learning based LCD framework

In this section, we will briefly introduce the BiGANs based adversarial learning based LCD approach shown in Figure 2. This framework is combined by four steps:
1) LiDAR projection maps extraction;
2) Adversarial feature learning with unlabeled maps;
3) Feature extraction for given frames;
4) Sequence matching.

### A. 2D top-view Extraction

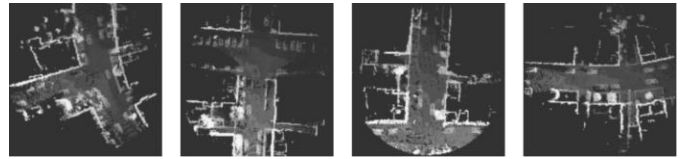

**Figure 3** 2D top-view maps

In the 2D top-view map extraction step, because of the low resolution of raw LiDAR scans, using single scan directly to generate projection maps may result in sparse map representation. Instead, we firstly use the octree map [20] with sequence LiDAR inputs to generated the local voxel map. Octree is a tree based data structure, and the occupancy of each leaf nodes in octree are updated by a log-odds method. For the detail about octomap, we recommend the reader to reference the original Octomap [20].

The map scale in Octomap is global scale and static, so the mapping efficiency will reduced with the map scale growing. In this paper, we only keep the octree nodes that within a given distance to the robot. Finally, the 2D top-view maps is generated by projected the local octree map on the ground plane as shown in Figure 3.

## B. BiGANs based Adversarial Feature Learning

As shown in Figure 1(b), BiGANs [17] is combined with three modules. A decoder module *De* aims to generate the synthesis data *G-X* from the low dimension random noise *Z*; an encoder module *En* transforms the real data *R-X* into the latent code domain *E-Z*; a discriminator module *D* is responsible for distinguishing the joint distribution of *(X, En(X))* and *(De(Z), Z)*. Same as in the original GANs [21], we could use a min-max value function to meet the above requirements,

$$\min_{De,En} \max_{D_J} V(D_J, De, En)$$
$$= E_{x \sim P_x}\left[\log D_J(x, En(x))\right]$$
$$+ E_{z \sim P_z}\left[\log(1 - D_J(De(z), z))\right] \quad (1)$$
$$= E_{x \sim P_x}\left[E_{z \sim P_{En}(\cdot|x)}\left[\log D_J(x, z)\right]\right]$$
$$+ E_{z \sim P_z}\left[E_{x \sim P_{De}(\cdot|z)} \log(1 - D_J(x, z))\right]$$

where $P_z$ is the random latent code distribution, $P_x$ is the real data distribution. As proved by the Donahue [17], with fixing the generator module (*En* and *De*), the optimal discriminator $D_J^*$ is the *Radon-Nikodym derivative*,

$$f_J := \frac{dP_{EX}}{d(P_{EX} + P_{GZ})} : \Omega \mapsto [0,1] \quad (2)$$

where $P_{EX}$ is the joint distribution of *(R-X, E-Z)* and $P_{GZ}$ is the joint distribution of *(G-X, Z)*. In this case, the value function $V(D_J^*, De, En)$ could be rewritten as,

$$C(E,G) = \max_D V(D, E, G) = V(D_{EG}^*, E, G)$$
$$= E_{(X,Z) \sim P_{EX}}\left[\log D_{EG}^*(x,z)\right] + E_{(X,Z) \sim P_{GZ}}\left[\log(1 - D_{EG}^*(x,z))\right]$$
$$= E_{(X,Z) \sim P_{EX}}\left[\log f_J(x,z)\right] + E_{(X,Z) \sim P_{GZ}}\left[\log(1 - f_J(x,z))\right]$$
$$= D_{KL}\left(P_{EX} \| (P_{EX} + P_{GZ})\right) + D_{KL}\left(P_{GZ} \| (P_{EX} + P_{GZ})\right) \quad (3)$$
$$= D_{KL}\left(P_{EX} \| \frac{P_{EX} + P_{GZ}}{2}\right) + D_{KL}\left(P_{GZ} \| \frac{P_{EX} + P_{GZ}}{2}\right) - \log 4$$
$$= 2D_{JS}(P_{EX} \| P_{GZ}) - 2\log 2,$$

where $D_{KL}$ is the Kullback-Leibler (KL) and $D_{JS}$ is the Jensen-Shannon divergences. Since JSD is always non-negative, so the value function could only reach its global optimal when $P_{EX} = P_{GZ}$. With the encode module, we could estimate the latent code from the real data.

## C. Feature Extration

The features from the 2D top-view maps is estimated by applying the forward operation in the adversarial feature learning networks. And this operation could be easily achieved by the embedded Jetson TX1 card, which is low power consumption and with a power built-in GPU supported. The above property enable our method to apply on the traditional robot system for the real time navigation task.

To measure the difference between frames, we simply use the Euclidean distance,

$$Diff(v_i, v_j) = \|v_i - v_j\|^2 \quad (4)$$

where $v_i$ is the encoded latent-code from the frames.

## D. Sequence Matching

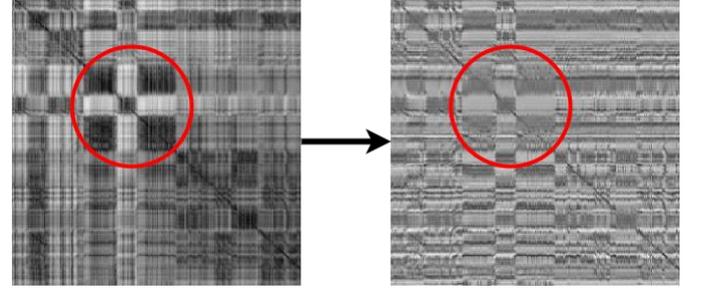

**Figure 4** Difference Matrix and Enhanced Matrix. Each pixel represents the feature similarity of relative test and pre-stored frame. The darker the pixel, the more similar of the relative single frame match. The red circles show the significant result of local enhancement.

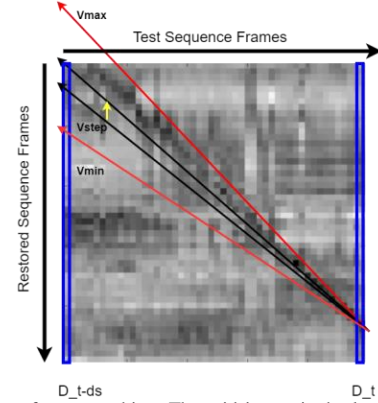

**Figure 5** Sequence frame matching. The grid image is the local difference matrix with a look back window length *ds*, $V_{min}$ and $V_{max}$ represent the minimum and maximum velocity for different score *S* calculation. $V_{step}$ is the step value for velocity selection. Finally, the best sequence match is estimated with the lowest *S*.

With the given test sequence frames and pre-stored frames, we could obtain the difference matrix by using the latent codes based on Equation 3, as shown in the left side of Figure 4. However, the matching scores in the difference matrix are highly related with the neighbor matches. And the most similar matches may stacked into a sub area. Since, sequence matching rely on the sum of differences in routes, matches stacking may reduce the LCD accuracy in sequence matching, as shown in the red circle of Figure 4. This problem could be solved by using an local enhancement operation [8],

$$\hat{D}_i = \frac{D_i - \bar{D}_l}{\sigma_l} \quad (5)$$

where $\bar{D}_l$ and $\sigma_l$ is the mean and standard deviation of the neighbor matches. Finally, to recognize the best matches, a space window M with the recent image difference vectors is used,

$$M = \left[\hat{D}^{T-d_s}, \hat{D}^{T-d_s+1} ... \hat{D}^T\right] \quad (6)$$

where $D^T$ is the column vector as shown in the right blue box in Figure 5, which represent the difference vector of test frame at timestamp $T$ with the pre-stored sequences. $d_s$ is the time length to watch back for sequence searching. $S$ is the sum of the frame difference of routes under different velocities,

$$S = \sum_{t=T-d_s}^{T} D_k^t, k = s - V(d_s - T + t) \quad (7)$$

where $s$ is the relative position in train frame, $V$ is the potential velocity proportion of test sequence and train sequence. The loop closures are then could be estimated when the difference score $S$ is small than a given threshold.

### III. STABLE ADVERSARAL FEATURE LEARNING

Before introduce the stable adversarial feature learning, we first investigate the instability in the BiGANs.

#### A. Instability Analysis

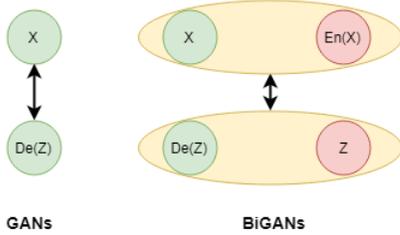

**Figure 6** Traction procedure for the GANs and BiGANs.

Figure 6 shows the traction procedure in the GANs and BiGANs. In GANs, the discriminator module is used to measure the distance between synthesis data distribution and real data distribution; the generator module is used for pulling the synthesis data distribution towards the target data distribution.

In BiGANs, the discriminator module needs to measure the distance between the joint distributions, which is a more complex manifold than in the GANs. As proved by Arjovsky et.al [19], the more complex manifold will lead to the synthesis distribution and target distribution hardly having measureable distance with the JSD. Such problem lead to uncertainty in the real distance measurement, and the gradient of loss function is equal to zero at almost anywhere..

In the LCD task, to obtain the unique description for the local scenes, the required latent codes should capture the geometry detail in the real data domain as much as possible. However, we could not easily achieve this goal with the BiGAN based joint distribution convergence. The goal in our work is to improve the joint distribution convergence, and make sure the latent codes could better represent geometry details.

#### B. Network Updating

Firstly, we use two approaches to improve the attractive force for joint distribution convergence: Wasserstein based joint loss function and cycle reconstruction as shown in Figure 7. Instead of using the original JSD to measure distances of the different distributions, we use the Wasserstein GAN (W-GAN) [22] proposed by Martin et.al. In an intuitive view, Wasserstein metric measure the total 'cost' to move one distribution to another. Given two distribution, measurements in Wasserstein is continuous. For the Jensen-Shannon (JS) divergence, the Total Variation (TV) distance, and Wasserstein distance, their metrics could be derived by,

$$W(P_\theta, P_0) = |\theta|,$$
$$JS(P_\theta, P_0) = \begin{cases} \log 2, if\ \theta \neq 0, \\ 0, if\ \theta = 0, \end{cases} \quad (8)$$
$$TV(P_\theta, P_0) = \begin{cases} 1, if\ \theta \neq 0, \\ 0, if\ \theta = 0. \end{cases}$$

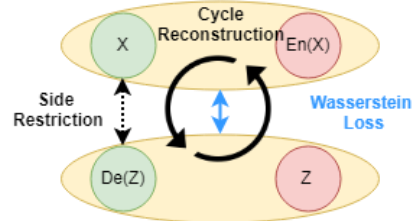

**Figure 7** Traction procedure for SAFL.

As shown in Figure 8, both JSD and TVD are all not continuous and could not provide stable gradient; while for the Wasserstein metric, the difference could maintain continuous even when the two distributions have no overlaps. In this paper, we skip the proving progress for transform Wasserstein distance into Wasserstein based loss function in W-GAN, for the detail place refer to Martin's work [22]. With the Wasserstein metric, the original value function of BiGAN in Equation 1 is then updated into the following one,

$$L_J(D_J, De, En) = \min_{De, En} \max_{\|D_J\|_L \leq 1} E_{x \sim P_{data}} \left[ D_J(x, En(x)) \right] - \quad (9)$$
$$E_{z \sim P_z} \left[ D_J(De(z), z) \right],$$

where $D_J$ is the discriminator for the joint distribution, and $||D_J||_L<=1$ indicates that we assume $D_J$ is locally Lipschitz. To achieve local Lipschitz, i.e. to have parameters in $D_J$ lie in a compact space, we adopt the simple approach in W-GAN, by clipping the parameters into a fixed box ([-0.01, 0.01]).

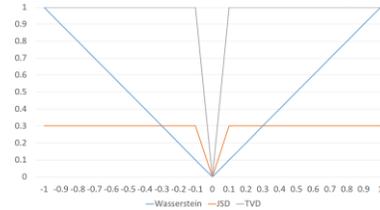

**Figure 8** Distance Measurement with different Metrics.

Since in the original BiGANs [17], the global optimal is achieved when $P_{EX}=P_{GZ}$, i.e. the two conditions X~De(En(X)) and Z~En(De(Z)) should be satisfied. To assist the attractive force between the joint distributions, we add the cycle reconstruction in both data and latent code domain with the $L2$ losses,

$$L_{CYC}(En, De) = \min_{En,De} E_{x \sim P_{data}} \left[ \|En(De(x)) - x\|^2 \right] + \\ E_{z \sim P_z} \left[ \|De(En(z)) - z\|^2 \right] \quad (10)$$

Secondly, to enhance the latent code could better capture the geometry details, we introduce a side updating module as shown the dash arrows in Figure 7. This GANs based adversarial loss are applied only for data domain, and the more the synthesis data looks like the real one, the more geometry details the latent codes could capture. For this additional side updating module, we use the original JSD loss to avoid the confliction with the joint distribution updating,

$$L_X(D_X, De) = \min_{De} \max_{D_X} E_{x \sim p_{data(x)}} \left[ \log D_X(x) \right] + \\ E_{z \sim p_z} \left[ \log(1 - D_X(De(z))) \right], \quad (11)$$

All the above module is organized as in Figure 9. Finally, combine the above loss functions, the full value function $V_{Joint}$ is obtained by,

$$De^*, En^*, D_X^*, D_Z^*, D_J^* = \\ \arg \min_{De, En} \max_{\|D_J\|_L \leq 1, D_X} (L_J + L_X + L_Z + L_{CYC}) \quad (12)$$

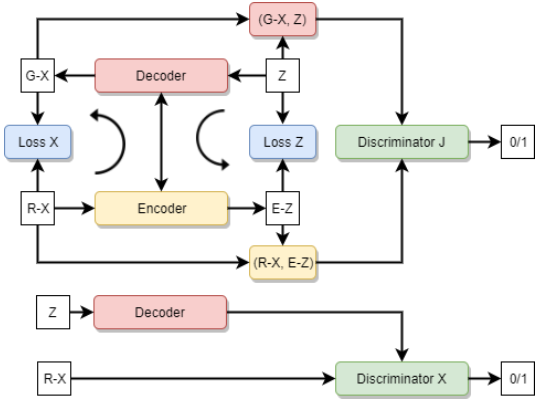

**Figure 9** The enhanced BiGAN framework.

## IV. EXPERIMENT AND ANALYSIS

To investigate the accuracy of our proposed method, we conduct the experiments with the KITTI odometry datasets. In this datasets, there are 22 LiDAR sequences and only sequences 00~10 have the ground truth of GPS location. We use sequence 01~08 to generate the 2D top-view maps for training, and use sequences 00, 09 and 11 for testing. The experiment is tested on the Ubuntu 14.04 system with a single NVidia Titan X card and 64G RAM supported. The 2D top-view map extraction is applied on the robot operation system (ROS).

Since the main source of the missing matches for the LiDAR based inputs is the viewpoints differences. The KITTI based LiDAR points, the local roll, pitch and higher differences are reduced in the octree mapping step. Thus the major source in viewpoints differences are the *Translation error* on the ground plane and the *Heading difference*. To test the robustness to the two transformation errors, we generate noise sequences $T\{T_x\}\_R\{R_y\}$ with 2D random noise (altitude is $T_X$ meters) in translation and 1D random noise (altitude is $T_X$ radian) in heading. Since our inputs is LiDAR based 2D top-view inputs, so it's meaningless to make the comparison with the new proposed Change-Removal [14] or other appearance based sequence matching methods [13] [15]. In this paper, we only make the comparison of our Stable-AFL method with the traditional sum of absolute differences (SAD) in the original SeqSLAM and the BiGAN features based the sequence matching framework. The parameters used in sequence matching are listed in table 2.

TABLE I
PARAMETERS USED IN ENHANCED SEQSLAM

| Parameter | Description |
|---|---|
| $d_s$ | The length of watch back trajectory, in this experiment, we set $d_s$=10 |
| $V_{min}$ | Minimum trajectory velocity proportion, here $V_{min}$=0.8 |
| $V_{max}$ | Minimum trajectory velocity proportion, here $V_{min}$=1.1 |
| $V_{step}$ | The step forward value for velocity proportion, $V_{step}$=0.1 |
| $D_{thresh}$ | The distance of matched points to decide whether matchings are satisfied, $D_{thresh}$=10m |

Here $d_s$, $V_{min}$, $V_{max}$, and $V_{step}$ are the relative parameters in Figure 5. And $D_{thresh}$ is the distance threshold to judge whether the matching are satisfied.

### A. Measurement Metrics

To measure the LCD accuracy of different methods, we make qualitative analysis with PRC (Precision-Recall curve) and AUC (area under the Receiver operating characteristic (ROC)); for the quantitative analysis, we use the recall at 100% perception in the PRC to measure LCD accuracy. Here, for the matched pairs, if the distance between ground truth position and estimated one is within $D_{thresh}$, then the pairs are regarded as true positive (TP), else will be regarded as false positive (FP); on the other side, the pairs erroneously discarded by the match score are regarded as false negative (FN), and the ones of actually no-matched pairs are regarded as the true negative (TN). Thus the precision and recall are then obtained by,

$$\text{Precision} = \frac{TP}{TP + FP} \\ \text{Recall} = \frac{TP}{TP + FN} \quad (13)$$

The AUC score is the size of covered ROC area, and the ROC curve is created by plotting the true positive rate (TPR) against the false positive rate (FPR) at various threshold settings, which are obtained by,

$$TPR = \frac{TP}{TP + FN}$$
$$FPR = \frac{FP}{TN + FP}$$
(14)

*B. Accurancy analysis*

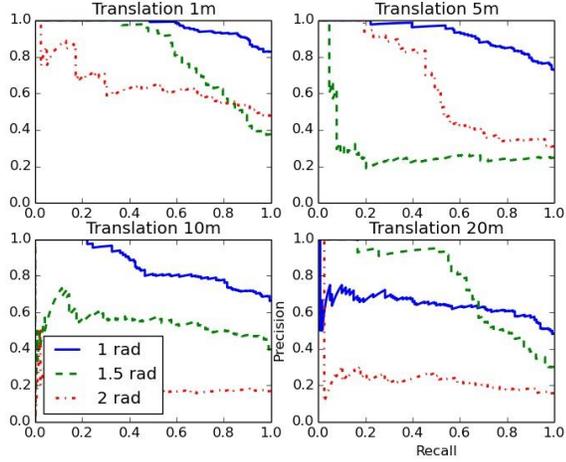

**Figure 10** Precision-Recall Curve for the original SeqSLAM method under different transformation errors

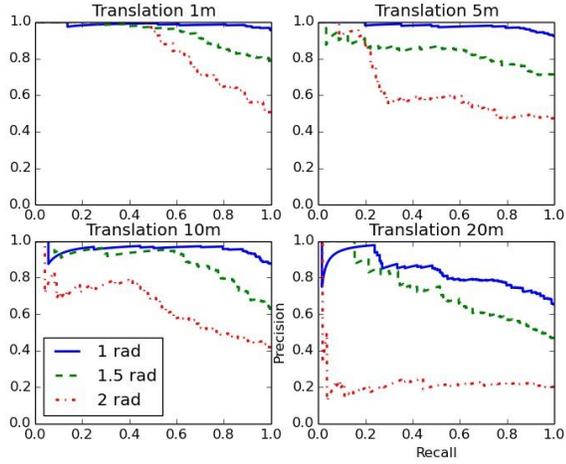

**Figure 11** Precision-Recall Curve of Original BiGAN based feature for sequence matching.

In this section, we will compare the LCD accuracy of different methods. For the qualitative analysis, Figure 10~12 give a PRC demonstration, and Figure 13 gives the AUC demonstration; for the quantitative analysis, Table II gives the recall at 100% precision.

Firstly, Figure 10 shows the PRC results based on the SAD features in the original SeqSLAM. In the cases of heading errors under 1 radian, SeqSLAM could still guarantee a stable LCD accuracy; while in the cases of higher heading error, the LCD accuracy reduces significantly.

Secondly, Figure 11 shows the results of original BiGAN based sequence matching, the LCD accuracy under different transformation errors are both improved, especially for the higher heading errors.

Thirdly, Figure 12 shows the results of our proposed Stable-AFL feature based sequence matching. Our method further improve the LCD accuracy in both lower and higher transformation errors. Especially for the cases of higher heading errors, our method could still achieve the LCD detection where other method fails.

For the more vivid demonstration on the LCD accuracy of different methods under variants transformation errors, Figure 13 shows the AUC index of ROC curve. As we can see from the index, our proposed Stable-ALI method is better than the SAD features or BiGAN features for the LCD detections.

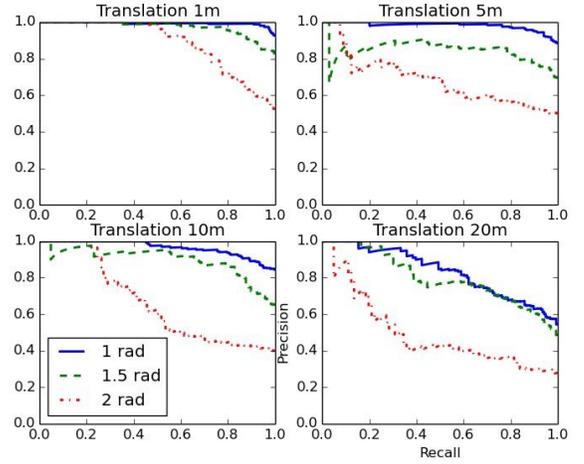

**Figure 12** Precision-Recall Curve of SLFL 2D feature based sequence matching under different noising sequences with ULFL features.

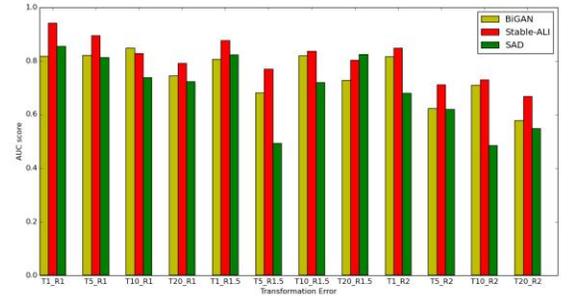

**Figure 13** AUC index of different methods under different transformation error. Where *T{a}_R{b}* represents that given the test sequence with random translation error within *a* meters and random rotation error within *(-b/2, b/2)* rad.

Table II gives the quantitative demonstration with the recall indexes under precision rate at 100%. In the T1_R1 case, SAD features achieve 44.5% recall, BiGAN based sequence matching achieve 51.9% recall, while our proposed method increases this to 90.9%, which is 204.6% and 175.1% times than the previous two methods. For the higher transformation errors, the Stable-ALI feature based sequence matching could still guarantee a stable LCD detection. The relative matching results could be found in our YouTube site[1].

---

[1] https://youtu.be/srOsTccVShA

TABLE II
RECALL AT 100% PRECISION

| Viewpoints difference | T1 R1 | T5 R1 | T10 R1 | T20 R1 | T1 R1.5 | T5 R1.5 | T10 R1.5 | T20 R1.5 | T1 R2 | T5 R2 | T10 R2 | T20 R2 |
|---|---|---|---|---|---|---|---|---|---|---|---|---|
| *SAD* | 28.3% | 44.5% | 12.4% | 0.8% | 1.3% | 0.8% | 0.9% | 2.1% | 0.2% | 0.1% | 0.0% | 0.0% |
| *BiGAN* | 51.9% | 19.8% | 35.7% | **19.8**% | 26.5% | 3.8% | 9.1% | 15.4% | 44.2% | **8.6**% | 4.2% | 1.8% |
| *Stable-ALI* | **90.9**% | **58.4**% | **46.1**% | 15.4% | **58.1**% | **12.5**% | **15.6**% | **16.5**% | **45.3**% | 7.5% | **10.8**% | **4.9**% |

TABLE IIII
FEATURE INFERENCE TIME PER FRAME (MILLISECOND)

| Viewpoints difference | T1 R1 | T5 R1 | T10 R1 | T20 R1 | T1 R1.5 | T5 R1.5 | T10 R1.5 | T20 R1.5 | T1 R2 | T5 R2 | T10 R2 | T20 R2 |
|---|---|---|---|---|---|---|---|---|---|---|---|---|
| *SAD* | 3.2 | 3.3 | 3.3 | 3.2 | 3.3 | 3.2 | 3.3 | 3.3 | 3.2 | 3.3 | 3.3 | 3.3 |
| *BiGAN* | 18.4 | 16.9 | 17.6 | 16.9 | 17.7 | 17.7 | 15.6 | 17.2 | 17.6 | 18.5 | 16.3 | 18.1 |
| *Stable-ALI* | 16.1 | 15.8 | 16.4 | 16.0 | 17.4 | 17.0 | 15.5 | 16.1 | 16.8 | 18.1 | 16.5 | 16.9 |

## C. Training analysis

In this section, we aim to compare the stability of the original BiGAN and our proposed stable-ALI. Here we use the AUC indexes to describe the LCD accuracy under different transformation errors, as shown in Figure 14. The x axis represents the epoch number of network models, y axis indicates the AUC index. We can see that the training process of our proposed method is more stable than the BiGAN method, and the AUC index of our proposed method could reach its optimal more fast than the BiGAN.

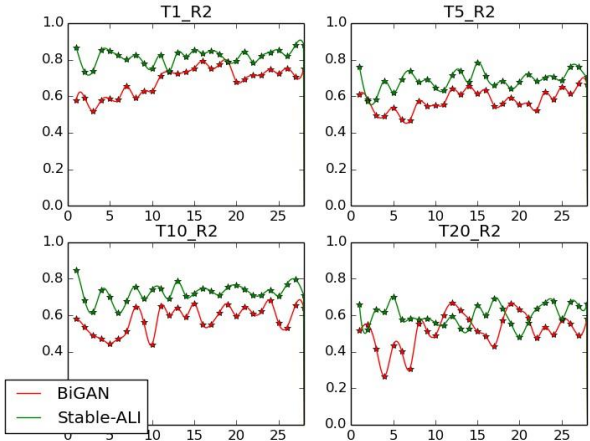

**Figure 14** AUC index under different transformation errors and different epoch of network models. X axis represents the epoch number of network models Y axis indicate the AUX index.

## D. Runtime and Storage analysis

For the runtime analysis of feature inference, the average feature inference time is shown in Table III. For our proposed method, the feature extraction could be done in 15~18ms per frame, which satisfies the real time requirement for normal robots navigation task. The features are saved as a 1024 vector in the float32 format, with the occupancy of 1024*4B=4KB. If we generate the code at 5Hz, after 24 hours running, the storage requirement for saving all the latent codes is only about 5*60*60*24*4KB~1.65GB. Thus the proposed method could be easily plugged on the any kinds of mobile robots for the real time long term navigation task, with relative small computation power and storage requirement.

## V. CONCLUSION

In this paper, we propose an unsupervised feature learning framework for the LiDAR projected 2D top-view map based LCD task. The 2D top-view LiDAR map is invariant to illumination changes, appearance changes, and viewpoints changes. We proposed a stable adversarial feature learning method, which is based on the original BiGAN, and combined with the cycle restriction module and the side updating module. The proposed method could enhance the attractive force between the joint distributions, and assist the efficient feature extraction for the LCD task. The experiments conducted on the KITTI odometry datasets shows that, our proposed method is better than the original SeqSLAM and BiGAN based sequence matching. In the case of translation error at 1 meter and heading error at 1 radian, the recall of enhanced BiGAN at 100% precision rate has increased 204.6% than SeqSLAM and 175.1% than BiGAN based sequence matching. In our future work, our will continue to investigate how to further improve the robustness for the viewpoints differences, and also try to add the latent codes from the visual inputs for better LCD detections.